\pdfoutput=1

\documentclass[11pt]{article}

\usepackage[]{acl}

\usepackage{times}
\usepackage{latexsym}

\usepackage[T1]{fontenc}

\usepackage[utf8]{inputenc}

\usepackage{microtype}

\usepackage{graphicx}
\usepackage{algorithm, algorithmic}
\usepackage{multirow}
\usepackage{caption}
\usepackage{subcaption}

\usepackage{tabularx}
\newcommand{\pipeline}{\textsc{StonyBook}}
\newcommand{\website}{\url{https://www.stonybook.org}}
\newcommand{\repo}{\url{https://github.com/sbu-dsl/stonybook}}

\title{\pipeline: A System and Resource for Large-Scale Analysis of Novels}

\author{Charuta Pethe, Allen Kim, Rajesh Prabhakar, Tanzir Pial, Steven Skiena \\ Department of Computer Science, \\ Stony Brook University, NY, USA \\ \texttt{\{cpethe,allekim,rprabhakar,tpial,skiena\}@cs.stonybrook.edu}}

\begin{document}
\maketitle
\begin{abstract}
Books have historically been the primary mechanism through which narratives are transmitted. 
We have developed a collection of resources for the large-scale analysis of novels, including:
(1) an open source end-to-end NLP analysis pipeline for the annotation of novels into a standard XML format, (2) a collection of 49,207 distinct cleaned and annotated novels, and (3) a database with an associated web interface for the large-scale aggregate analysis of these literary works.
We describe the major functionalities provided in the annotation system along with their utilities.  We present samples of analysis artifacts from our website, such as visualizations of character occurrences and interactions, similar books, representative vocabulary, part of speech statistics, and readability metrics. We also describe the use of the annotated format in qualitative and quantitative analysis across large corpora of novels. \footnote{Please check out the demo video at \\ \url{https://youtu.be/kCu-zKFvvQE}}.
\end{abstract}

\section{Introduction}
Books have historically been the primary mechanism through which knowledge and narratives are transmitted through time and space.  
Books (particularly novels) are a uniquely interesting class of texts that provide a view of the inner lives of people, what they think and how they work: knowledge that cannot be readily gleaned from other sources. Building intelligent systems requires that machines understand people on some fundamental level -- their goals, dreams, and thought processes.    Computational analysis of novels provides a possible approach to getting this understanding into intelligent computer systems, with wide-spread implications for common-sense reasoning and human factors engineering.

But the analysis of longform texts has generally been understudied by the natural language processing (NLP) research community.
Modern neural network techniques for NLP generally focus on short texts (such as paragraphs or social media posts) because model capacity limitations preclude the effective understanding of full longform texts.
We believe that improved data and computational resources will help open this field to a broader research community.

The field of digital literary studies has spurred advances in computational linguistics as well as digital humanities. However, the lack of standard annotation systems and exploratory interfaces for the large-scale computational analysis of books has hindered the efforts of these communities. This paper describes our book analysis system \pipeline\footnote{\repo} and associated web resource \footnote{\website} that aims to bridge this gap.

\begin{figure*}[t]
    \centering
    \includegraphics[width=0.85\linewidth]{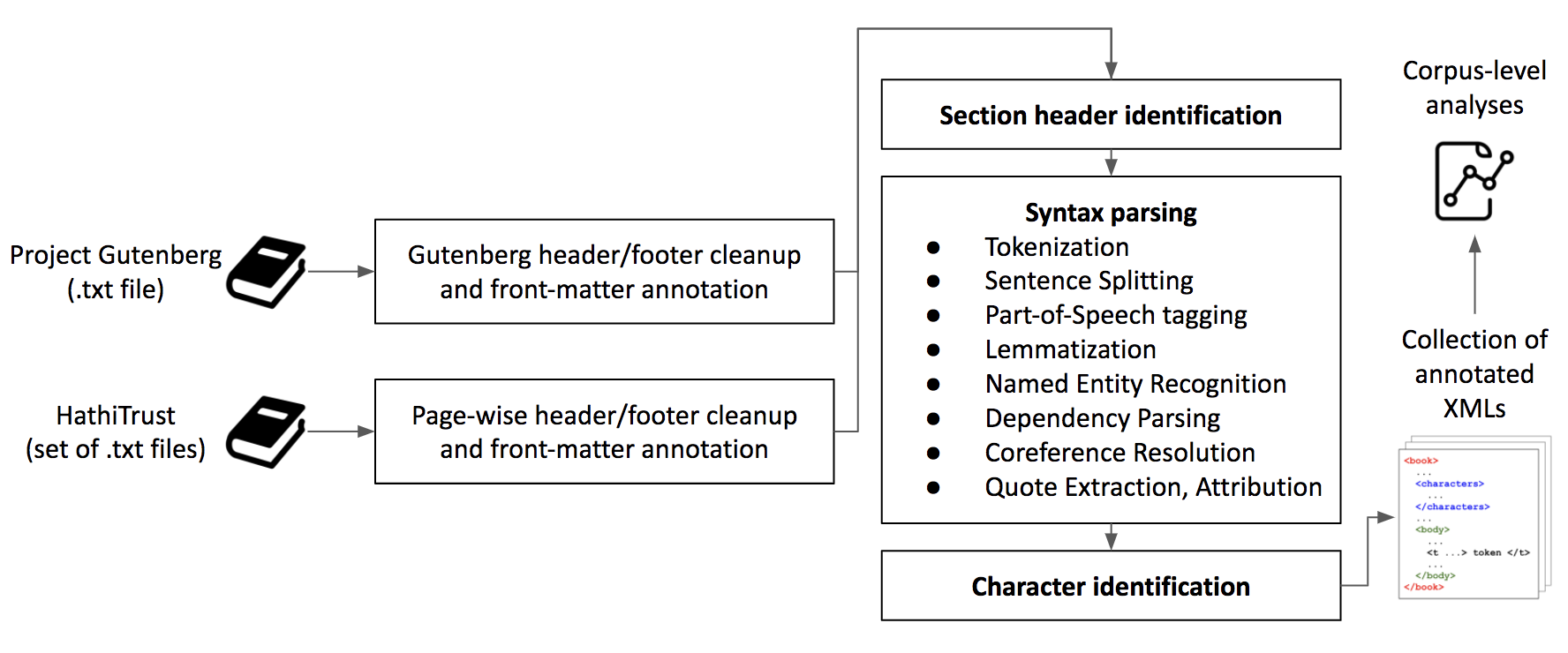}
    \caption{\pipeline: System overview}
    \label{fig:pipeline}
\end{figure*}

Our main contributions in this paper are:

\begin{itemize}
    \item An \textbf{end-to-end system} (in the form of a Python library) for the NLP-driven annotation of novels in a standard XML specification format, supporting inputs for different book file formats (Project Gutenberg text files, HathiTrust page-wise text files) with extensive text-cleaning efforts.  We describe the capability of our pipeline, including named entity recognition, quote extraction/attribution, and character identification.
    
    \item A \textbf{data resource} of 17,289 annotated books from the Project Gutenberg corpus and 31,918 other distinct annotated books from the HathiTrust Digital Library corpus. Beyond just annotating texts, we have performed extensive volume deduplication and the assembly of metadata including authorship year and current publication status.
    
    \item An \textbf{exploratory web interface} that provides statistics and insights about books, authors, and subjects, and where they lie in context of other books in the corpora.   As examples of the power of this resource, we present interesting observations on character frequency distributions and part of speech distributions.
\end{itemize}

Figure \ref{fig:pipeline} shows an overview of the system and the contributions, described in detail in the following sections. Due to copyright restrictions, we cannot directly distribute annotated texts from the HathiTrust dataset. However, the list of unique IDs for books from both corpora (Project Gutenberg and HathiTrust) along with analysis artifacts have been made available publicly on the website, facilitating a wide range of analyses.


\section{Related Work}
\label{sec:relatedwork}

GutenTag \cite{brooke-etal-2015-gutentag} is a tool that provides functionality to build subcorpora from the Project Gutenberg corpus, and a general tagging framework in order to facilitate interaction between the digital humanities and computational linguistics communities. The Gutenberg module of NLTK \cite{bird-loper-2004-nltk} provides a limited set of functionalities for a few books in the corpus.

The BookNLP pipeline \cite{bamman-etal-2014-bayesian} generates NLP-driven book analyses in multiple formats, with tokenization, part-of-speech tagging, dependency tagging, character identification, and quote attribution. This pipeline uses CoreNLP \cite{manning-etal-2014-stanford} for syntax parsing. BookNLP defines a set of initial characters and creates a set of allowable variants for each one, similar to \newcite{davis2003methods} and \newcite{elson-etal-2010-extracting}, after which each character is greedily assigned to the most recently linked entity of which it is a variant.

\newcite{groza2015information} use an ontology-based approach to identify characters, that is specific to folktales. \newcite{vala-etal-2015-mr} present a graph-based approach to cluster variants of character name mentions, involving heuristics to decide whether to merge two vertices. The CHARLES tool \cite{vala-etal-2016-annotating} provides a collaborative interface for manual annotation of characters. \newcite{dekker2019evaluating} compare four different approaches for character identification, by evaluating against a manually annotated dataset. \newcite{jahan-finlayson-2019-character} propose a refined definition of character, and an approach that uses an animacy detector followed by a supervised classification model to label which of the animated entities qualify as characters. \newcite{egbert2020fiction} focus on analysis of register variation in fiction.

\section{System Components}
\label{sec:architecture}

In this section, we describe the primary data sets we employ, and the analysis phases we use to clean and process these into annotated texts. This work is focused on the effort and discipline with which we unify disparate corpora, tools, and metadata to create a substantial resource for research on novels.

\subsection{Datasets}
\label{sec:datasets}

Our primary resources are two well-known collections of scanned texts:

\begin{itemize}
\item
{\em \textbf{Project Gutenberg}} --
We identified 17,289 books by 4,425 authors from the English language section of the Project Gutenberg corpus, tagged with the term "fiction" as subject. Previous research, such as \citet{raecompressive2019}, utilized Project Gutenberg extensively and introduced PG-19, a book-level language modeling benchmark dataset derived from it. The StonyBook pipeline can aid researchers working with Project Gutenberg books.

\item
{\em \textbf{HathiTrust Digital Library}} --
We identified a subset of 31,918 books from 14,496 authors from the \newcite{hathitrust_dataset} dataset, distinct from the Project Gutenberg dataset. We obtained this subset by filtering from a set of English fiction books \cite{underwood_dataset}. 

\end{itemize}

We followed \newcite{kim-etal-2021-cleaning-dirty} for removing duplicate copies of books from both the datasets based on title, author, and content similarity.

\subsection{Annotation Format}

After each of the phases described in the subsequent sections, we store the text along with its annotations in XML format, adding or removing tags as required after each phase. At the end of the pipeline, we have an XML file conforming to a standard specification, consistent across corpora, simplifying our analysis. The final annotated file for each book includes information about POS, NER, dependency, and character counts. An example can be found in Figure \ref{fig:char_xml}.

To process these XML files, we provide utility functions developed using lxml \cite{behnel2005lxml} that we use to generate our analysis. This will be especially useful to those who prefer to store the annotations in a different format (e.g. collection of CSV files as done in BookNLP).

\subsection{Preprocessing}
To annotate Gutenberg headers and footers, we employ an improved version of the cleanup module from the Gutenberg Python package\footnote{\url{https://pypi.org/project/Gutenberg/}}. To annotate front and back matter in books from the HathiTrust corpus, we use a neural classification model \cite{mcconnaughey-etal-2017-labeled}.

\subsection{Section Header Identification}
We use an exhaustive rule-based search \cite{pethe-etal-2020-chapter} to identify and tag section headers. This covers section headers including key words like "Chapter", "Book", etc. We also factor in the chapter numbering to ensure consistency in our tagging. 


\subsection{Syntax Parsing}
We obtain the following annotations, as described in Figure \ref{fig:pipeline}: tokenization, sentence splitting, part-of-speech tagging, lemmatization, NER, dependency parsing, coreference resolution, and quote extraction and attribution. We provide the option to use either module for annotation: CoreNLP through Stanza \cite{qi-etal-2020-stanza} or spaCy with neuralcoref \cite{spacy2}.




\begin{figure}[htbp]
    \centering
    \includegraphics[width=\linewidth]{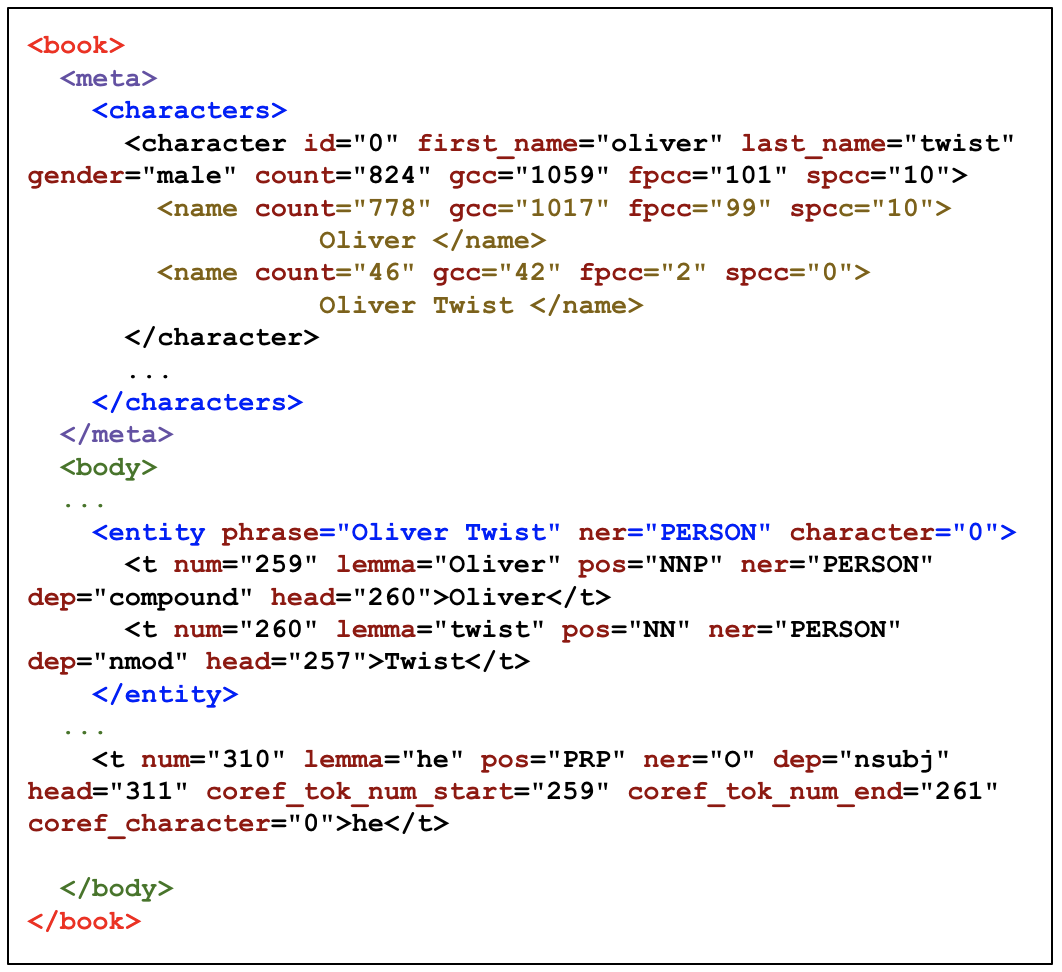}
    \caption{Processed XML including character information (\texttt{gcc}, \texttt{fpcc}, \texttt{spcc} stand for gendered, first-person and second-person coreference counts respectively)}
    \label{fig:char_xml}
\end{figure}
\subsection{Character Identification}
We perform the following steps to canonically identify the characters in a book:

\paragraph{Candidate augmentation: } We expand the \texttt{ner PERSON} tags to include any prefixed honorifics, and tag all occurrences of the entity as candidates.

\paragraph{Gender inference:} We infer the character's gender using three methods: dictionary lookup of the honorific, pronoun coreference majority vote, and an external data source \footnote{Gender Guesser (\url{https://pypi.org/project/gender-guesser/})}.

\paragraph{Name mention clustering: } For single name mentions, we map to the closest occurrence of a full name with either the same first name or last name, in the order of preference for gender inferred from honorific, coreferences, and name. We then find unique combinations of first/last names and gender. We sort these by frequency, and keep only those characters with more than 2 total mentions, to obtain the final character list.

Figure \ref{fig:char_xml} shows a sample template of the processed XML after character identification. This includes information about character names, genders, occurrences in the text, and types of coreferences and their counts.

\section{Web Interface}
\label{sec:web_interface}

The results and metadata of analysis of each of almost 50,000 novels appears on our website \website.
Our website features detailed analysis of each book, described in this section. Additionally, we provide an author-wise and subject-wise view of all these novels. For easy perusal of our website, we provide a list of links to analyses of popular books. This list of books was procured using ratings and popularity data from GoodReads \footnote{\url{https://www.goodreads.com/}} and Amazon \footnote{\url{https://www.amazon.com/}}.

\subsection{Characters}

Unifying and aggregating character references into distinct character personas enables us to measure the importance and interactions between characters, and how this evolves over the length of the book.

\begin{figure}[htbp]
    \centering
    \includegraphics[width=\linewidth]{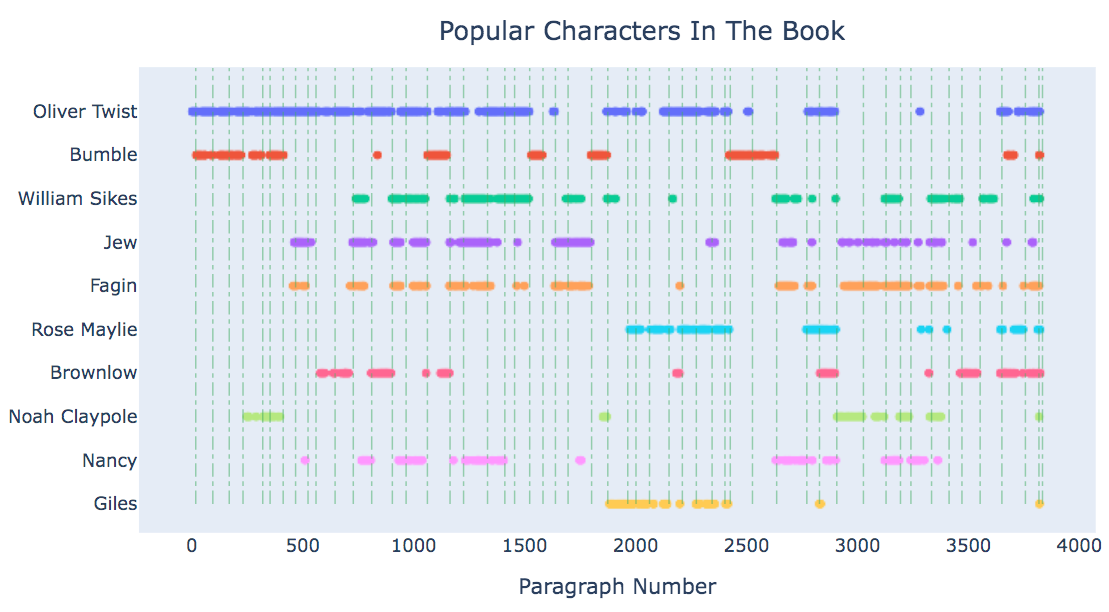}
    \caption{Popular characters and their occurrences in \textit{`Oliver Twist'} (vertical dashed lines denote chapter breaks).}
    \label{fig:char_dots}
\end{figure}

Figure \ref{fig:char_dots} shows a part of the exploratory web interface for the book \textit{`Oliver Twist'} by Charles Dickens. It shows where each of the top 10 most frequent characters occur throughout the narrative. This gives us an idea of the positional importance of characters, which has potential applications in generating character representations, and book-level QA involving questions about specific characters.

\begin{figure}[h]
    \centering
    \includegraphics[width=0.7\linewidth]{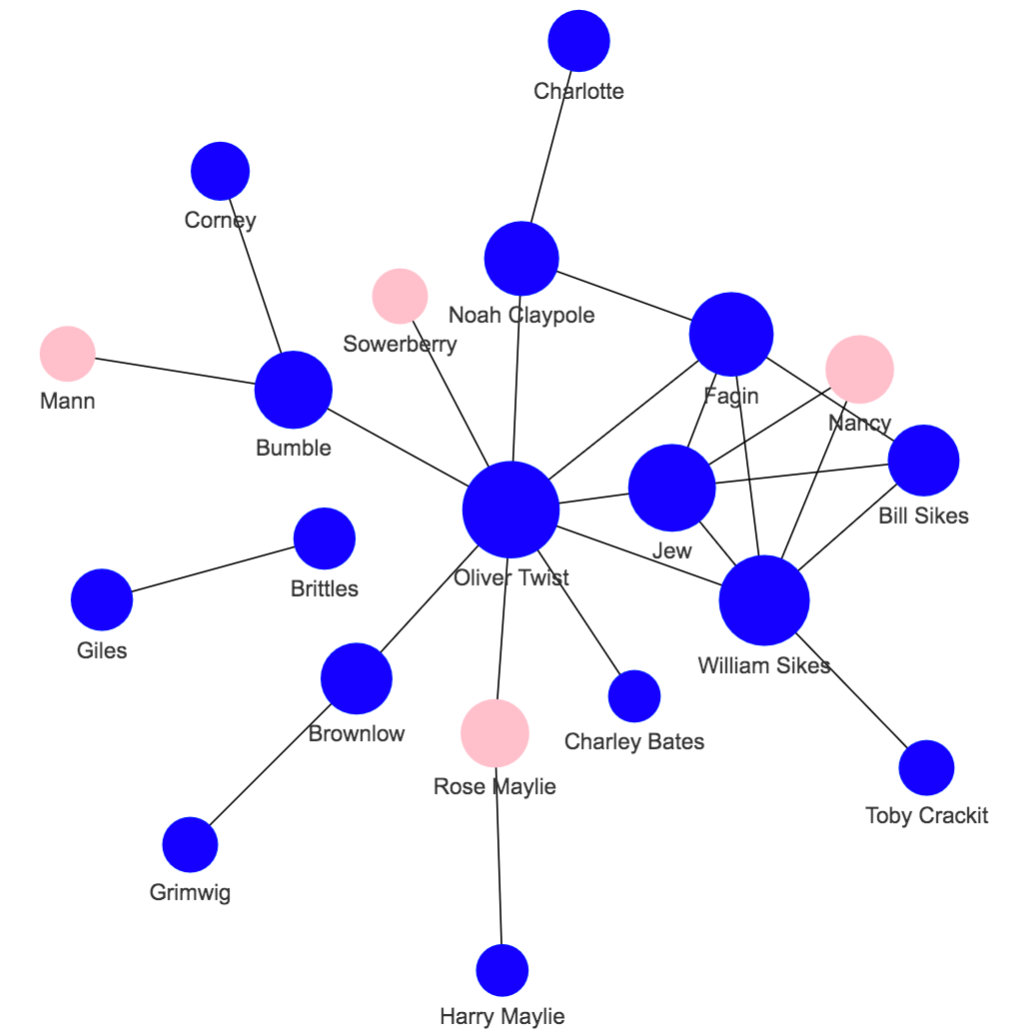}
    \caption{Character interaction network for \textit{`Oliver Twist'} (Blue nodes represent male characters, pink nodes represent female characters, node size is proportional to the number of occurrences of the character).}
    \label{fig:char_graph}
\end{figure}

We also extract a character interaction network from the book, where we consider two characters to be frequently interacting if they occur within a fixed window (here we empirically set the window size to 30 tokens, but this is a modifiable parameter) of each other more than 5 times. We display this network in the web interface, as shown in Figure \ref{fig:char_graph}, along with character occurrences.

\begin{figure}[htbp]
    \centering
    \includegraphics[width=\linewidth]{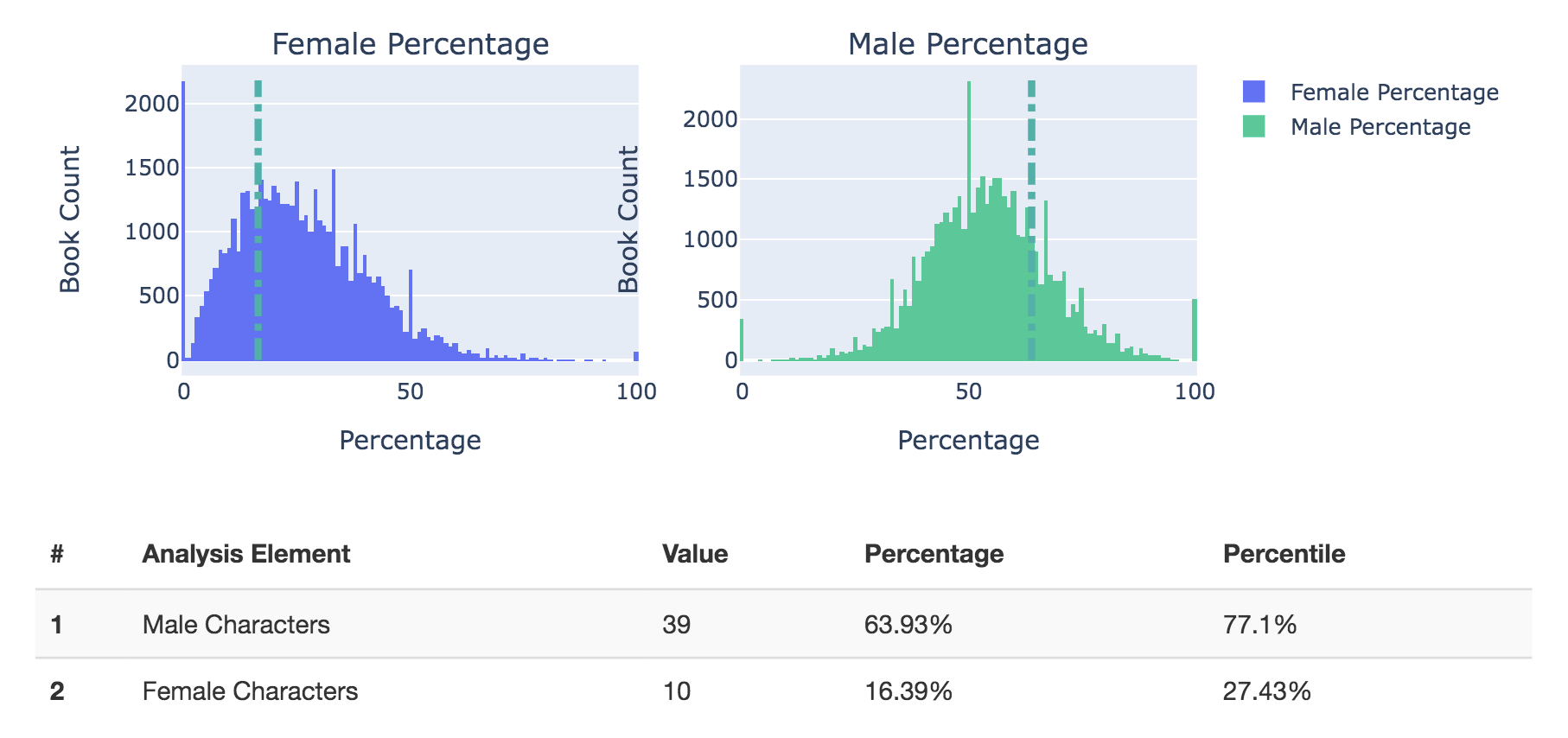}
    \caption{Gender statistics for \textit{`Oliver Twist'}.}
    \label{fig:gender_stats}
\end{figure}

We also display the distribution of the percentage of male and female characters across the entire corpus, and denote where each book lies in the distribution, as shown in Figure \ref{fig:gender_stats}.

In addition to these visualizations, we also display a comprehensive list of characters in the book, with all their aliases, mention counts, and pronoun coreference counts.

\subsection{Most Similar Books}

\begin{figure}[htbp]
    \centering
    \includegraphics[width=0.9\linewidth]{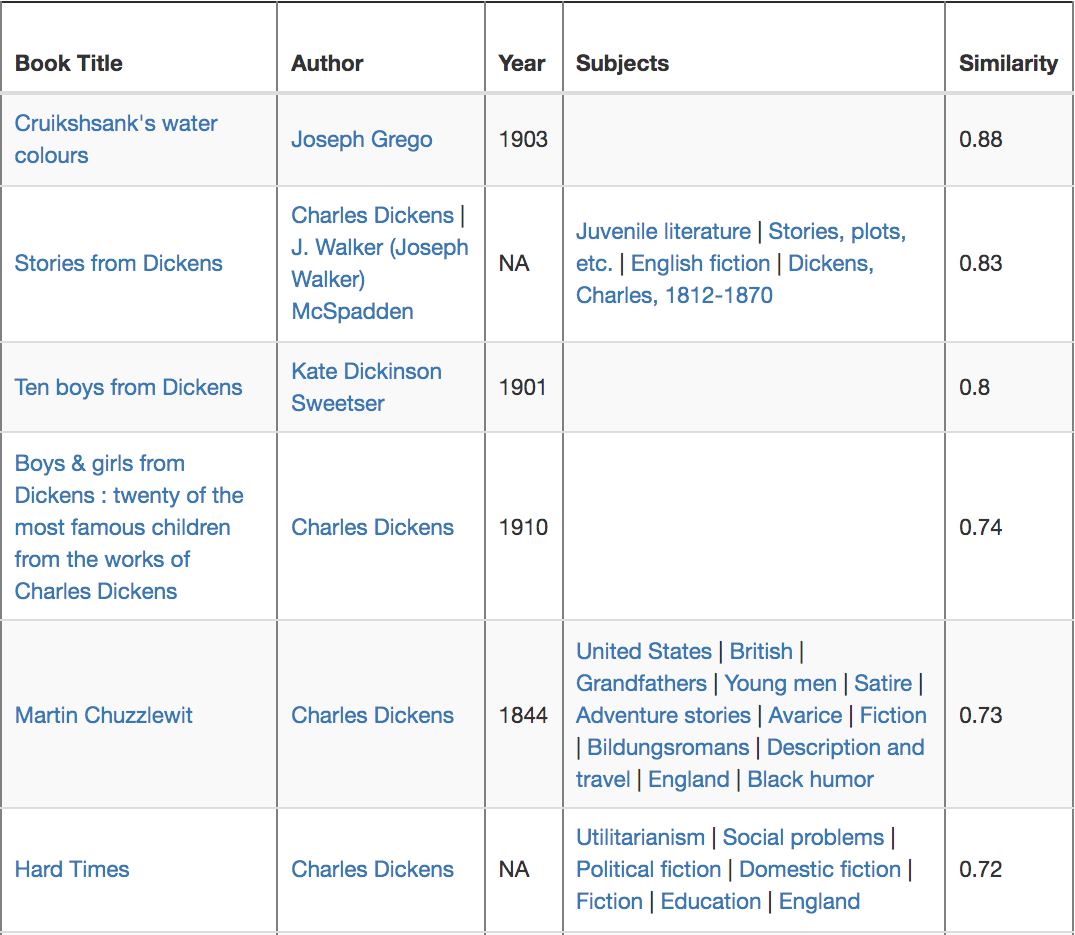}
    \caption{Top most similar books for \textit{`Oliver Twist'}.}
    \label{fig:similar_books}
\end{figure}

Book embeddings provide vector representations facilitating queries of the similarities of different works.
We train a doc2vec model \cite{le2014distributed} on all the books in both the corpora (Project Gutenberg and HathiTrust) combined. We train for 10 epochs using a window size of 5, to generate 100 dimensional vectors. We restrict our vocabulary to the top 200,000 words and exclude words with counts less than 100. We strip stop words and lemmatize each token. Then for each book, we list the top 10 most similar books from each of the corpora as shown in Figure \ref{fig:similar_books}.

\subsection{Representative Vocabulary}

\begin{figure}[h]
    \centering
    \includegraphics[width=0.8\linewidth]{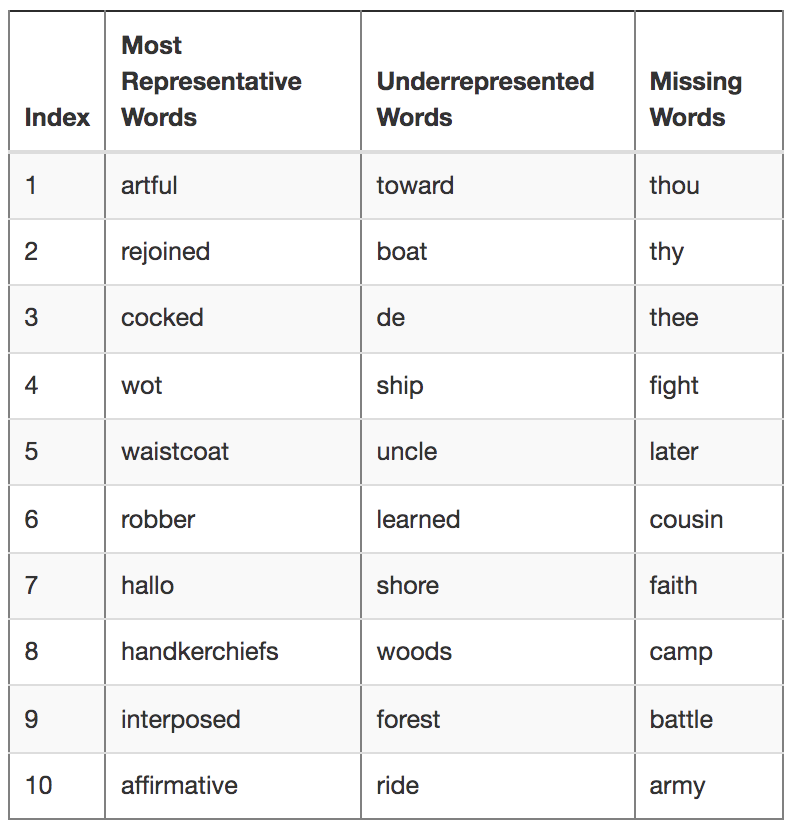}
    \caption{Representative vocabulary for \textit{`Oliver Twist'}.}
    \label{fig:rep_vocab}
\end{figure}

For each book, we compute the ratio of the normalized frequency of each word, and compute the ratio of this with the globally normalized word frequency across all books. We sort words by this ratio to obtain lists of most and least representative words in the book, filtered to the 10,000 most common words in the corpus. We also list the words that are most common in the corpus but are absent in the book, as missing words. Figure \ref{fig:rep_vocab} shows a part of our web interface that shows representative vocabulary.

\subsection{Part of Speech Statistics}

\begin{figure}[htbp]
    \centering
    \includegraphics[width=\linewidth]{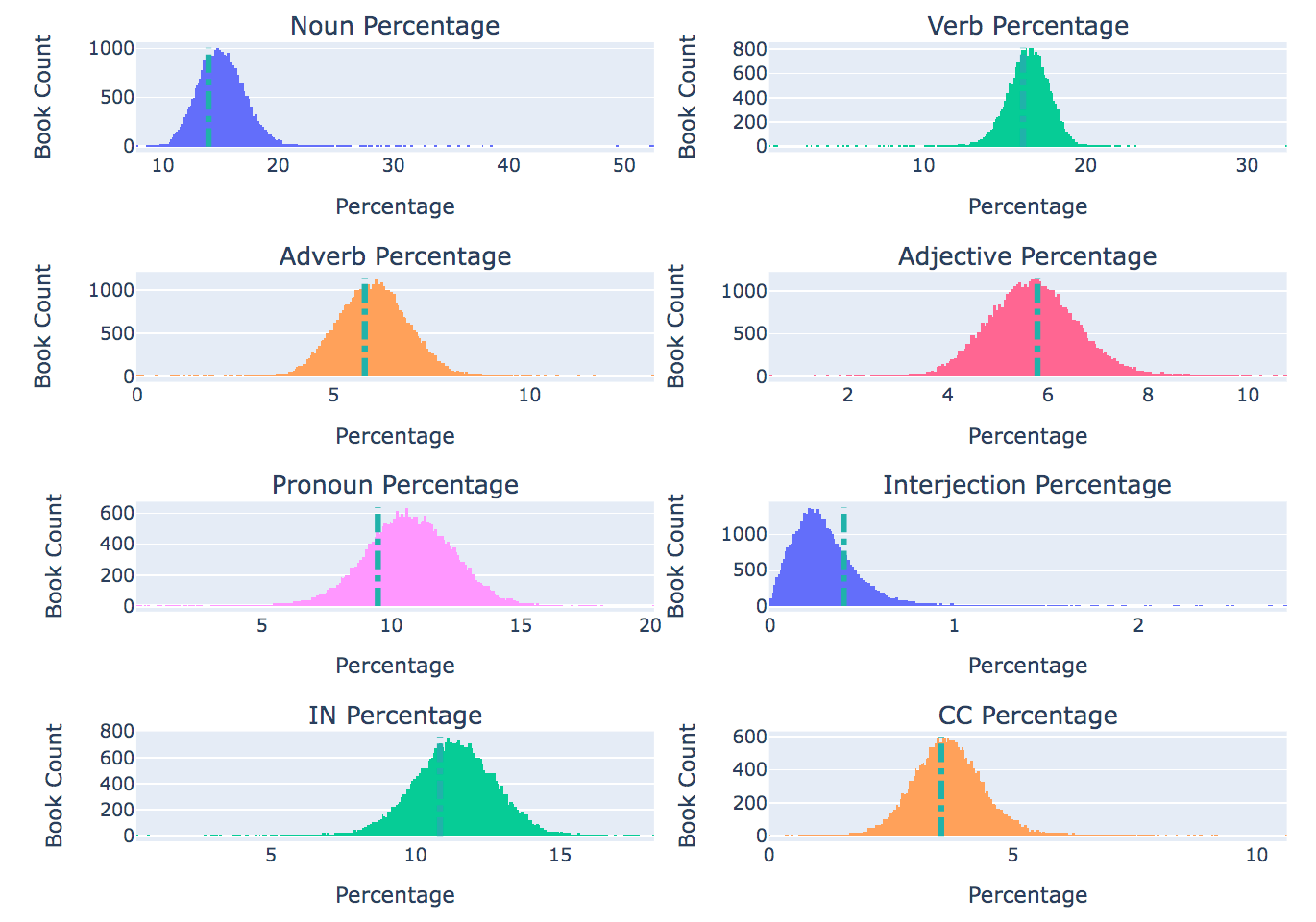}
    \caption{Part of speech distribution graphs for \textit{`Oliver Twist'}.}
    \label{fig:pos_distrib}
\end{figure}

For each of the following parts of speech (POS), we compute the count and percentage among all POS tags within the book, as well as the percentile among all the books: Noun, Adjective, Verb, Adverb, Pronoun, Interjection, Preposition, and Conjunction. Additionally, as shown in Figure \ref{fig:pos_distrib}, we also display the distribution of POS percentages across all books in the corpus, and denote where each particular book lies in the distribution.

\subsection{Readability Metrics}

\begin{figure}[htbp]
    \centering
    \includegraphics[width=0.8\linewidth]{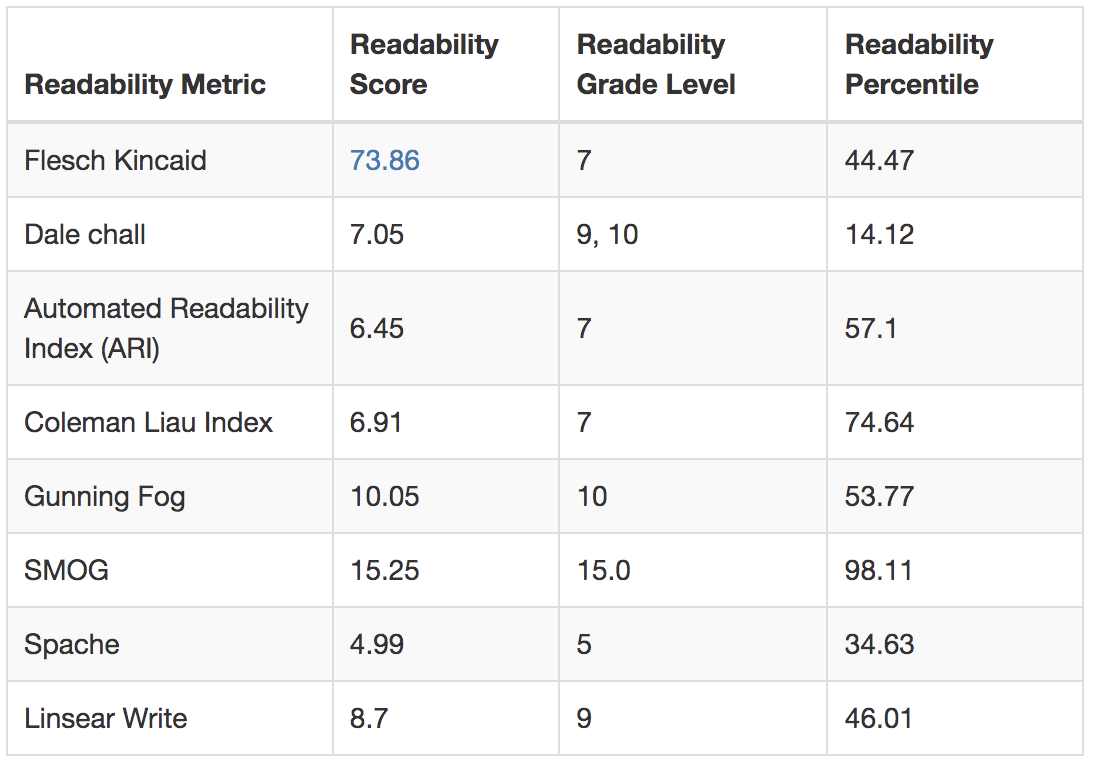}
    \caption{Readability metrics for \textit{`Oliver Twist'}.}
    \label{fig:readability}
\end{figure}

We compute the following readability metrics for each book: Flesch Reading Ease, Dale Chall Readability Score \cite{dale1948formula}, Automated Readability Index (ARI), Coleman-Liau Index \cite{coleman1975computer}, Gunning Fog Index \cite{kincaid1975derivation}, SMOG Index \cite{mc1969smog}, Spache Readability Score and Linsear Write Formula Grade level.

\section{Exploring the Resource}

\label{sec:insights}

In this section, we present some of the aggregate analyses powered from the dataset.
These should be viewed as explorations and demonstrations more than definitive treatments of these issues, in the hopes that they inspire further work.

\begin{figure}[htbp]
    \centering
    \includegraphics[width=0.8\linewidth]{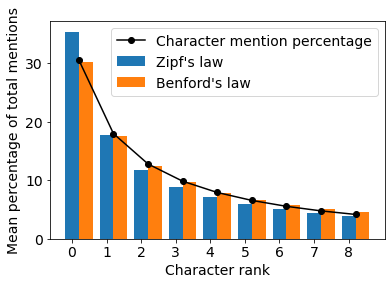}
    \caption{Mean percentage of the number of mentions of the top 9 most frequent characters for each character rank (across 45,939 books with at least 9 characters), with Benford's law and Zipf's law distributions plotted for comparison}
    \label{fig:char_rank_perc}
\end{figure}

\paragraph{Percentage frequency vs. character rank: } We consider all books in our combined corpus (Project Gutenberg + HathiTrust) that contain at least nine distinct characters. We rank characters in decreasing order of frequency, and for each rank, we compute the average percentage of this rank's mention across all books. Figure \ref{fig:char_rank_perc} shows the resulting plot, which reveals that the distribution of the percentage of character mentions as a function of rank is similar to the Zipf's law and Benford's law distributions \cite{benford1938law}.

\paragraph{Ratio of top 2 characters' mentions: } For all the books in the combined corpus, we compute the ratio of number of mentions of the most frequent character to the second most frequent character. For a large majority of books, this ratio is less than 10. However, the remaining books, i.e. the outliers reveal an interesting insight: Books with unusually high ratios feature a main protagonist who dominates most of the narrative. 
A notable example of an outlier is {\em Robinson Crusoe, told to the children} (67.75), understandably because it is a tale of a man castaway alone on a deserted island.

\begin{figure}[htbp]
    \centering
    \includegraphics[width=0.8\linewidth]{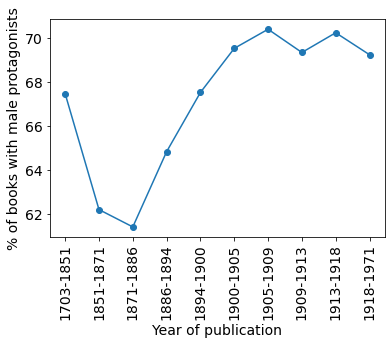}
    \caption{Percentage of books with male protagonists by year of publication}
    \label{fig:male_protagonist_perc}
\end{figure}

\paragraph{Protagonist's gender over time: } For books from the HathiTrust dataset, the year of publication is available as metadata. We sort these books by year of publication and divide them into ten equal subsets. Figure \ref{fig:male_protagonist_perc} shows the percentage of books with male protagonists in each subset, as a function of time. The period from 1851 to 1886 shows a sharp decline in the percentage of male protagonists. This time period features many books from prominent female authors, including Margaret Oliphant, Martha Finley, and May Agnes Fleming.

\begin{figure}[htbp]
    \centering
    \includegraphics[width=0.8\linewidth]{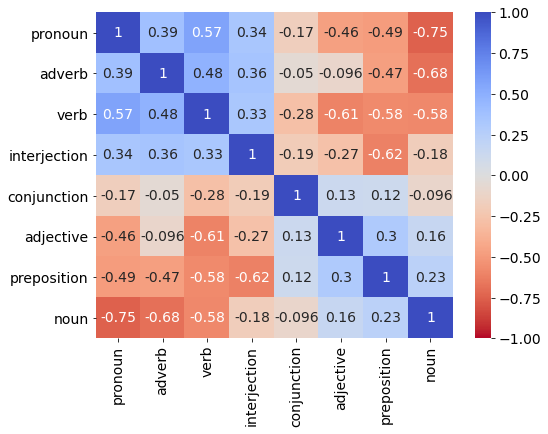}
    \caption{Pairwise correlations of POS tags: The parts of speech form two distinct groups inversely correlated with each other.}
    \label{fig:pairwise_pos_corr}
\end{figure}

\paragraph{POS Correlations: } Figure \ref{fig:pairwise_pos_corr} shows the pairwise Pearson correlations of eight parts of speech. Interestingly, the parts of speech form two distinct groups inversely correlated with each other. Intuitively, noun-adjective and verb-adverb are pairs with positive correlations. However, these pairs are in different groups, indicating that books lean towards being either more action-oriented or description-oriented.

\section{Conclusion}
\label{sec:conclusion}
We present an end-to-end system for NLP-driven annotation of literary works, along with a standard XML annotation format. Additionally, we also present a resource of annotated books in this standard format, along with analysis. This work will bridge the gap between computational analysis and linguistic exploration, and enable easier processing of books for further downstream tasks such as social network extraction, character embedding generation, event/action prediction, knowledge graph embedding, and causality analysis.



\bibliography{anthology,custom}
\bibliographystyle{acl_natbib}

\appendix



\end{document}